\documentclass[UTF8,11pt,a4paper]{ctexart}

\usepackage[a4paper,margin=2.5cm]{geometry}
\usepackage{amsmath,amssymb}
\usepackage{booktabs}
\usepackage{graphicx}
\usepackage{caption}
\usepackage{subcaption}
\usepackage{multirow}
\usepackage{threeparttable}
\usepackage[hidelinks]{hyperref}
\usepackage{enumitem}
\usepackage[super]{gbt7714}
\usepackage{xeCJK}
\begin{document}

\begin{center}
{\Large\bfseries 面向绿色价值评价的智慧农业平台环境效益仿真评估\\
基于蒙特卡洛方法的“减药—减肥—节水—降碳”不确定量化}\\[6pt]
{\normalsize（English Title: Monte Carlo-Based Ex-Ante Assessment of the Green Benefits of an AI-Driven Smart Agriculture Platform in Hainan）}\\[6pt]
{\normalsize 李肇阳，张锐杰，孙兆吉，张璐\\[2pt]
三亚学院（Sanya University），海南三亚 572022；邯郸未至之境人工智能基础软件有限责任公司}\\[2pt]
{\small 2026年9月}
\end{center}

\begin{quote}
\noindent\textbf{摘要：}为评价智慧农业决策平台在农药减量、化肥减量、灌溉节水和碳减排方面的绿色价值与环境效益，本文以“宏秋智农”AI智慧农业决策平台（集成大语言模型问答、病虫害智能诊断、物联网传感器、NASA GIBS卫星遥感与地块农事数据闭环）为对象，以海南芒果园、冬季瓜菜基地和水稻/南繁育种田三类代表性热带场景为试点基线，构建“投入品—田间排放”摇篮到田门（cradle-to-farm-gate）的农业碳排放核算模型，并将平台干预措施转化为可量化的传导参数，采用蒙特卡洛方法对参数不确定性进行量化。结果表明：在平台被试点农户充分采纳的情景下，化学农药使用量预计减少23.5\%（90\%置信区间15.0\%--33.2\%），化肥施用量减少21.0\%（13.8\%--28.9\%），综合灌溉用水节约16.5\%（10.9\%--23.5\%），单位面积碳排放强度下降21.5\%（16.1\%--27.2\%）；“化肥减量15\%以上”与“碳强度明显下降”两个目标达成概率较高（分别为90.6\%与98.1\%），而“综合节水20\%以上”达成概率仅约20\%，建议按场景区别表述。全局敏感性分析（Sobol一阶指数）表明，测土配方施肥比例与有机替代系数解释了综合碳强度下降方差的主要部分（合计约83\%），稻田CH$_4$减排在水稻场景中贡献显著。收敛性检验显示10\,000次迭代即可使统计量稳定，三情景分析给出保守—基准—乐观的取值范围。研究可为智慧农业绿色效益的量化评估与试点观测方案的指标优先序设计提供方法参考。
\end{quote}

\noindent\textbf{关键词：}智慧农业；绿色价值；蒙特卡洛模拟；碳足迹；农药化肥减量；不确定性分析

\newpage

\title{Monte Carlo-Based Ex-Ante Assessment of the Green Benefits of an AI-Driven Smart Agriculture Platform in Hainan}
\author{Zhaoyang Li, Ruijie Zhang, Zhaoji Sun, Lu Zhang\\
\small \textit{Sanya University, Sanya, Hainan 572022, China;}\\
\small \textit{Handan Weizhi Jingjie AI Basic Software Co., Ltd.}}
\date{September 2026}
\maketitle

\begin{abstract}
Smart agriculture platforms are widely regarded as key carriers for implementing China's pesticide/fertilizer reduction, water-saving and carbon-reduction agendas, yet a unified quantitative framework for assessing their green value is still lacking. Taking an AI-driven decision platform for tropical agriculture as the object (integrating large-language-model question answering, multimodal pest diagnosis, IoT soil-moisture sensing, NASA GIBS satellite remote sensing, and a closed-loop field record system), this study constructs a cradle-to-farm-gate agricultural carbon accounting model covering pesticide and fertilizer production, field N$_2$O, irrigation electricity and paddy CH$_4$, translates platform interventions into quantifiable transmission parameters, and propagates parameter uncertainty by Monte Carlo simulation over three representative Hainan scenarios (mango orchards, winter vegetables and rice/nanfan breeding fields, weighted 40\%:30\%:30\% by area). Under a fully-adopted pilot scenario, the median reductions are 23.5\% (90\% interval 15.0\%--33.2\%) for chemical pesticide use, 21.0\% (13.8\%--28.9\%) for fertilizer application, 16.5\% (10.9\%--23.5\%) for aggregate irrigation water, and 21.5\% (16.1\%--27.2\%) for carbon intensity. The probabilities of achieving fertilizer reduction $\geq$15\% and a clear carbon-intensity decline are high (90.6\% and 98.1\%, respectively), whereas the probability of aggregate water saving $\geq$20\% is only about 20\%, suggesting that scenario-specific statements are preferable. Sobol first-order sensitivity analysis shows that the soil-test-based fertilizer recommendation rate and the organic-substitution coefficient jointly explain about 83\% of the variance of aggregate carbon-intensity reduction, with paddy CH$_4$ mitigation contributing markedly in the rice scenario. Convergence tests show that 10,000 iterations stabilize all statistics, and conservative/baseline/optimistic scenario bounds are reported. The framework provides a reproducible, calibration-ready methodology for ex-ante green value assessment and for prioritizing pilot observation indicators.
\end{abstract}


\section{Introduction}

Agricultural green development is a key lever for China's rural revitalization and dual-carbon goals. Since 2024, the Rural Revitalization Plan (2024--2027) and the National Smart Agriculture Action Plan (2024--2028) have emphasized fertilizer and pesticide reduction, non-point source pollution control, and farmland quality protection; chemical input reduction, water-saving irrigation and paddy methane abatement have also entered the greenhouse-gas mitigation policy agenda. Against this background, smart agriculture platforms supported by artificial intelligence, the Internet of Things (IoT) and satellite remote sensing are regarded as key carriers for translating "pesticide reduction, fertilizer reduction, water saving and carbon reduction" into field-level practice \cite{moa2024,ccg2025}.

However, a unified quantitative framework for assessing the environmental benefits of smart agriculture platforms is still lacking. Most studies focus on single technical interventions, e.g., soil-testing-based fertilizer recommendation \cite{zhang2018}, integrated water-fertilizer management \cite{moa2013}, or water management for paddy methane mitigation \cite{IPCC2019,lindau1991}; few evaluate the whole platform integrating diagnosis--medication--irrigation--fertilization--remote sensing. Moreover, because field experiments are slow and costly, precision-agriculture carbon assessments usually rely on life-cycle analysis (LCA) and expert judgment with substantial parameter uncertainty, and many studies report single point estimates without intervals \cite{lam2014}.

Monte Carlo simulation assigns probability distributions to input parameters and propagates uncertainty under repeated random sampling, providing probability intervals and sensitivity rankings; it is a well-established tool for agricultural systems under parametric uncertainty \cite{saltelli2008}. Using tropical agriculture in Hainan as the background, this paper builds a Monte Carlo framework for the green value of a smart agriculture platform and answers three questions: (1) what are the attainable intervals of the four categories of green benefits ("pesticide reduction--fertilizer reduction--water saving--carbon reduction"); (2) what are the attainment probabilities of the platform's green targets (pesticide $\geq$20\%, fertilizer $\geq$15\%, water saving $\geq$20\%) at different confidence levels; (3) which parameters contribute most to output uncertainty and should be prioritized in pilot observation.

\section{Materials and Methods}

\subsection{Object of Study and Scenarios}

The "Hongqiu Zhinong" platform integrates five modules: an agricultural LLM question-answering assistant, multimodal pest diagnosis, an IoT sensor engine (soil moisture, temperature, EC, pH, etc.), multi-source satellite remote sensing analysis (mainly public NASA GIBS imagery), and a closed-loop field-operation record system. Its green-value transmission paths are fourfold: \textbf{(1) pesticide reduction}---intelligent diagnosis lowers the share of blind spraying, and medication records plus pre-harvest interval reminders reduce repeated and over-dosage applications; \textbf{(2) fertilizer reduction}---soil-testing-based prescription advice matched to soil and crop growth stage, on-demand topdressing, supplemented by organic substitution; \textbf{(3) water saving}---sensor-driven drip irrigation/integrated water-fertilizer management and paddy smart irrigation scheduling; \textbf{(4) carbon reduction}---transmission of the above reductions to input production and field emissions, plus paddy CH$_4$ mitigation by intermittent irrigation.

Three representative pilot scenarios are selected (Table~\ref{tab:scenario}): Hainan mango orchards (mu-year), winter vegetable bases (represented by cowpea rotation, mu-crop) and rice/nanfan breeding fields (mu-season). The aggregate indicator is area-weighted 40\%:30\%:30\%.

\begin{table}[htbp]
\centering
\caption{Pilot scenario definitions and platform interventions vs.\ conventional management}\label{tab:scenario}
\small
\begin{threeparttable}
\setlength{\tabcolsep}{4pt}
\begin{tabular}{lp{3.0cm}p{3.2cm}p{3.0cm}}
\toprule
Scenario & Unit & Platform intervention & Conventional management\\
\midrule
Mango orchard & mu-year & diagnosis, records, drip \& remote-sensing advice & calendar spraying, flood irrigation\\
Winter vegetable & mu-crop & diagnosis, records, interval reminders & calendar spraying, experience-based fert.\\
Rice/nanfan & mu-season & smart irrigation, water-layer decisions & continuous flooding, experience-based fert.\\
\bottomrule
\end{tabular}
\end{threeparttable}
\end{table}

\subsection{Transmission Model}

Platform interventions propagate to each indicator through the chain below (subscripts $m,v,r$ denote mango, vegetable and rice; $\Delta$ is the reduction rate relative to conventional management):

\begin{align}
\Delta_{pest} &= \varphi_a \times \eta_{rx}\\[2pt]
\Delta_{fert} &= 0.7\,\varphi_s + 0.3\,\varphi_o\\[2pt]
\Delta_{w,m} &= \eta_{d,m}\,\alpha,\quad
\Delta_{w,v} = \eta_{d,v}\,\alpha,\quad
\Delta_{w,r} = \eta_{i,r}\,\alpha\\[2pt]
\Delta_{w,agg} &= 0.4\,\Delta_{w,m}+0.3\,\Delta_{w,v}+0.3\,\Delta_{w,r}
\end{align}

where $\varphi_a$ is the share of avoidable blind sprays in conventional management; $\eta_{rx}=0.75$ is the prescription uptake factor (platform advice is not fully implemented; a conservative discount is applied); $\varphi_s$ is the reduction achievable by soil-testing recommendations; $\varphi_o$ is the organic-substitution coefficient (weight 0.3); $\eta_{d,m}$, $\eta_{d,v}$ and $\eta_{i,r}$ are the water-saving rates of mango drip irrigation, vegetable drip irrigation and rice smart irrigation, respectively; and $\alpha$ is the technology adoption rate.

\subsection{Carbon Accounting Scope}

Using a cradle-to-farm-gate scope, five components are accounted (kg CO$_2$e/mu-period): pesticide production, fertilizer production plus field N$_2$O, irrigation electricity, paddy CH$_4$, and machinery diesel:

\begin{equation}
C_{after} = E_{pest}(1-\Delta_{pest}) + E_{fert}(1-\Delta_{fert}) + E_{irr}(1-\Delta_{w}) + E_{ch4}(1-\Delta_{ch4}) + E_{mach}
\end{equation}

\begin{equation}
C_{base} = \sum_{j} E_j
\end{equation}

Paddy CH$_4$ is included only in the rice scenario, using the IPCC default continuous-flood factor with water-management reduction coefficients \cite{IPCC2006,IPCC2019}; field N$_2$O is estimated as 1\% of applied nitrogen directly emitted \cite{IPCC2006}; the irrigation electricity factor uses the national average grid emission factor (0.5366 kg CO$_2$/kWh, 2022 basis) \cite{mee2024}.

Baseline component values and their source grades are given in Table~\ref{tab:params}. Source grading: A = official document or international methodology; B = peer-reviewed literature; C = expert judgment (uncertainty expressed as an interval when data are missing).

\begin{table}[htbp]
\centering
\caption{Model input parameters, probability distributions and sources}\label{tab:params}
\small
\begin{threeparttable}
\begin{tabular}{llcll}
\toprule
Parameter & Distribution & Mode & Range & Grade\\
\midrule
Blind-spray share $\varphi_a$ & triangular & 0.30 & 0.15--0.50 & C\\
Soil-test reduction $\varphi_s$ & triangular & 0.15 & 0.05--0.30 & B\\
Organic substitution $\varphi_o$ & triangular & 0.30 & 0.10--0.55 & A/B\\
Technology adoption $\alpha$ & triangular & 0.55 & 0.30--0.80 & C\\
Mango drip saving $\eta_{d,m}$ & triangular & 0.35 & 0.20--0.55 & A\\
Vegetable drip saving $\eta_{d,v}$ & triangular & 0.30 & 0.15--0.50 & A\\
Rice smart-irrigation saving $\eta_{i,r}$ & triangular & 0.20 & 0.08--0.35 & B\\
Paddy CH$_4$ reduction $\Delta_{ch4}$ & triangular & 0.30 & 0.10--0.55 & B\\
Fertilizer C baseline mango (kg CO$_2$e/mu) & triangular & 105 & 80--140 & B/C\\
Fertilizer C baseline vegetable & triangular & 150 & 115--195 & B/C\\
Fertilizer C baseline rice & triangular & 145 & 110--185 & B/C\\
Paddy CH$_4$ C baseline rice (kg CO$_2$e/mu) & triangular & 148 & 95--230 & A\\
\bottomrule
\end{tabular}
\begin{tablenotes}\footnotesize
\item[Note] A = official documents/international methodologies (Ministry of Agriculture integrated water-fertilizer guidance \cite{moa2013}; IPCC guidelines \cite{IPCC2006,IPCC2019}; MEE emission factors \cite{mee2024}); B = peer-reviewed literature; C = expert judgment.
\end{tablenotes}
\end{threeparttable}
\end{table}

The remaining components (pesticide production, irrigation electricity, machinery carbon baselines) are scenario-specific triangular distributions (means about 18/22/8, 8/10/22 and 6/8/18 kg CO$_2$e/mu-period for mango/vegetable/rice), to be calibrated and replaced by pilot-collected data.

\subsection{Monte Carlo Simulation and Sensitivity Analysis}

Simulation uses fixed random seed 20260906 with 10,000 iterations, producing probability distributions of the four indicators. Global sensitivity analysis applies Sobol first-order indices estimated by Pick-and-Freeze \cite{saltelli2008}, with 8,000 samples per independent group:

\begin{equation}
S_{1,i} = \frac{\mathrm{E}\left[f(\mathbf{B})\left(f(\mathbf{A}^{B_i})-f(\mathbf{A})\right)\right]}{\mathrm{Var}(f)}
\end{equation}

In addition: (1) convergence tests (statistic drift over sample sizes 10$^2$--10$^5$); (2) three-scenario analysis (parameter endpoints: conservative = lower bound, baseline = mode, optimistic = upper bound); (3) comparison with literature-reported values. Computations are run in a Python environment with reproducible scripts.

\section{Results and Analysis}

\subsection{Probability Intervals of the Four Green Benefits}

Statistics from 10,000 iterations are reported in Table~\ref{tab:results}. The median chemical pesticide reduction is 23.5\% (90\% interval 15.0\%--33.2\%), fertilizer reduction 21.0\% (13.8\%--28.9\%); aggregate irrigation water saving 16.5\% (10.9\%--23.5\%); aggregate carbon-intensity reduction 21.5\% (16.1\%--27.2\%), with P5 positive (16.1\%), indicating robust direction.

\begin{table}[htbp]
\centering
\caption{Monte Carlo results (N=10$^{4}$, fixed seed 20260906)}\label{tab:results}
\small
\begin{threeparttable}
\begin{tabular}{lcccccc}
\toprule
Indicator & Scenario & Mean & SD & P5 & Median & P95\\
\midrule
Pesticide reduction & all scenarios & 0.238 & 0.055 & 0.150 & \textbf{0.235} & 0.332\\
Fertilizer reduction & all scenarios & 0.211 & 0.046 & 0.138 & \textbf{0.210} & 0.289\\
Water saving & mango & 0.202 & 0.055 & 0.120 & 0.197 & 0.300\\
 & winter vegetable & 0.174 & 0.051 & 0.098 & 0.169 & 0.267\\
 & rice/nanfan & 0.115 & 0.038 & 0.060 & 0.112 & 0.184\\
 & aggregate & 0.168 & 0.038 & 0.109 & \textbf{0.165} & 0.235\\
Carbon-intensity reduction & mango & 0.205 & 0.036 & 0.147 & 0.204 & 0.265\\
 & winter vegetable & 0.203 & 0.037 & 0.145 & 0.203 & 0.265\\
 & rice/nanfan & 0.242 & 0.046 & 0.167 & 0.241 & 0.319\\
 & aggregate & 0.215 & 0.033 & 0.161 & \textbf{0.215} & 0.272\\
\bottomrule
\end{tabular}
\end{threeparttable}
\end{table}

\begin{figure}[htbp]
\centering
\includegraphics[width=0.92\textwidth]{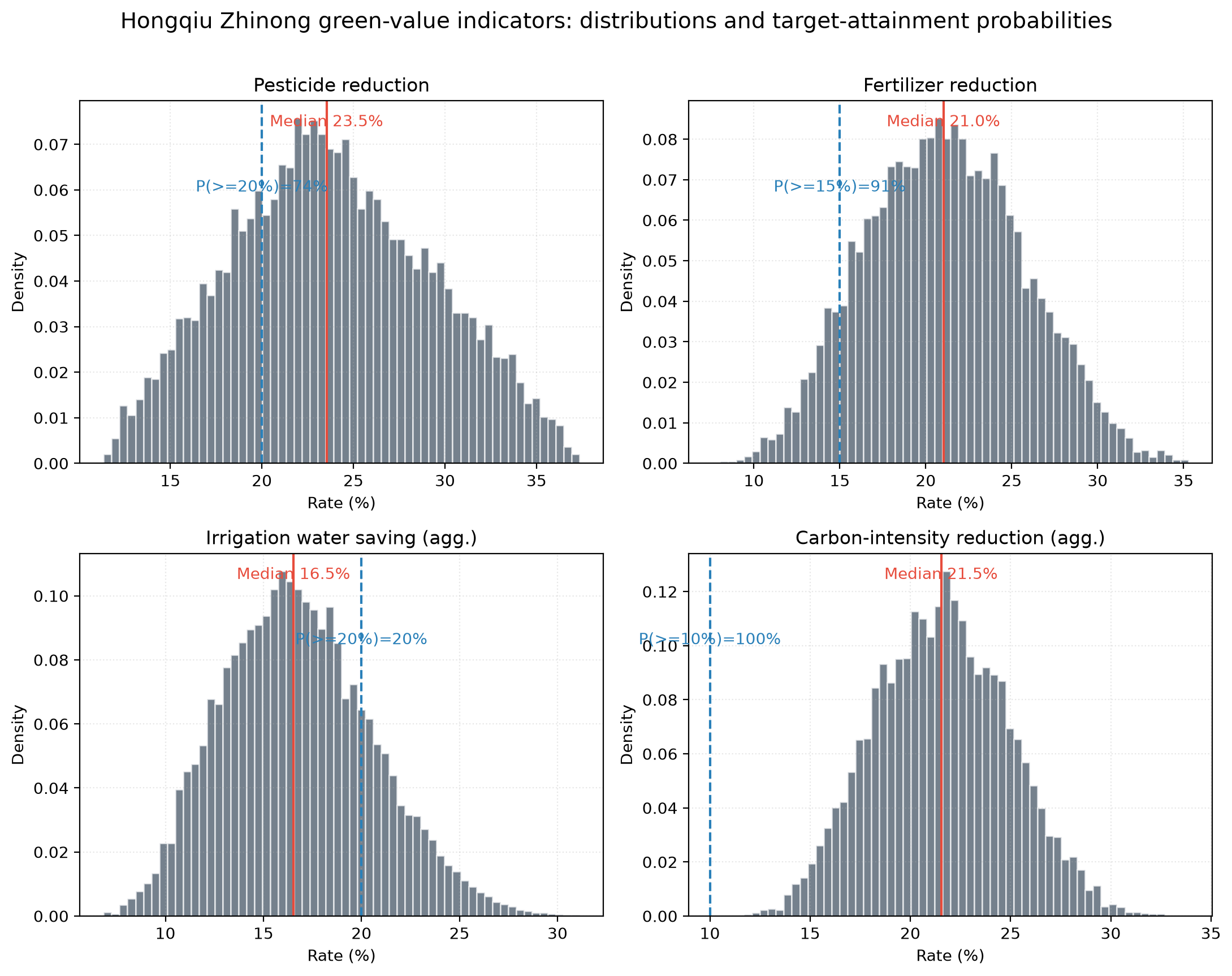}
\caption{Distributions of the four green-value indicators and target-attainment probabilities. Dashed lines are target reference values (pesticide 20\%, fertilizer 15\%, water 20\%, carbon intensity 15\%)}
\label{fig:dist}
\end{figure}

\begin{figure}[htbp]
\centering
\includegraphics[width=0.95\textwidth]{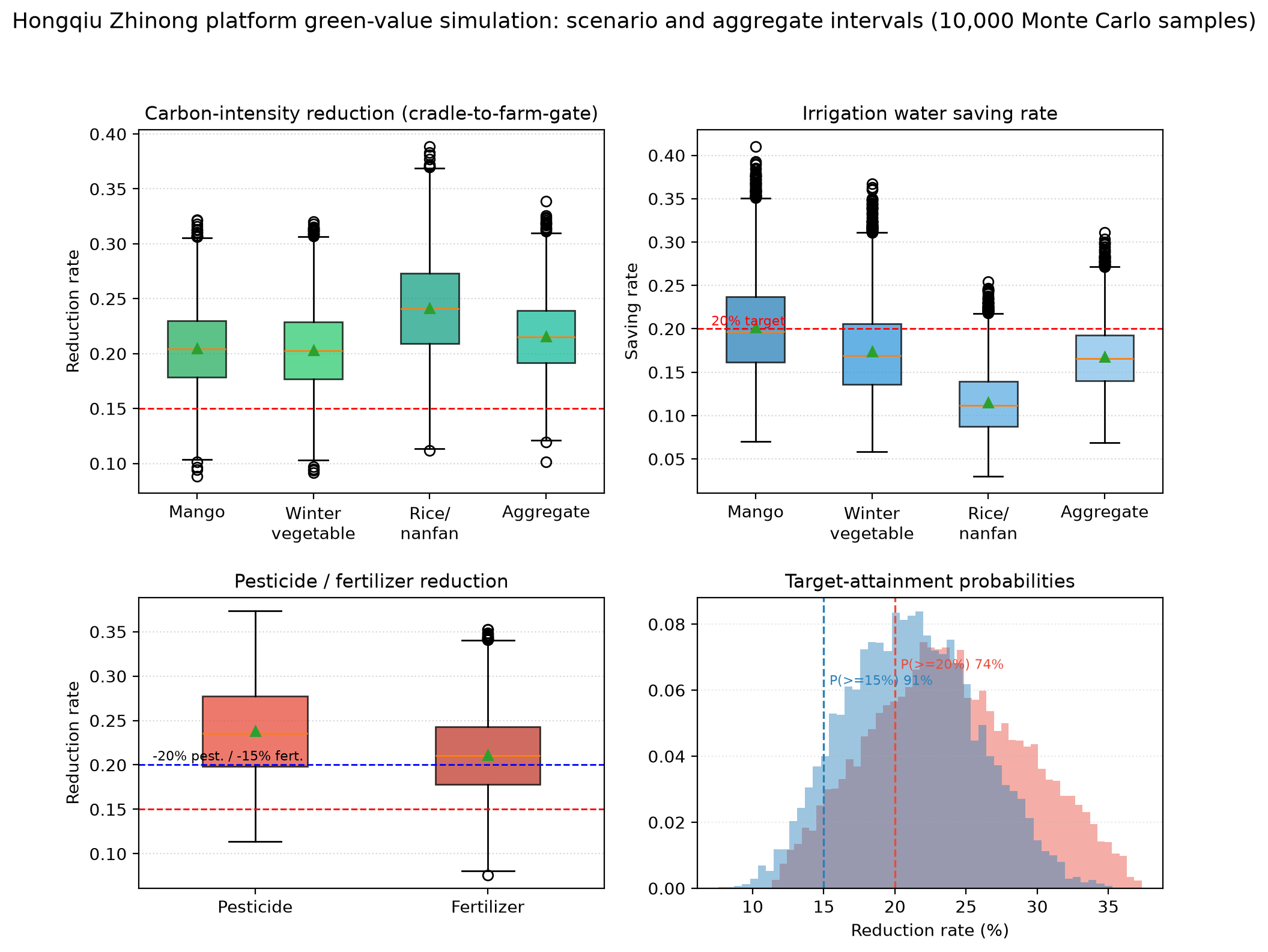}
\caption{Scenario boxplots of carbon-intensity reduction and irrigation water saving (10,000 MC samples)}
\label{fig:box}
\end{figure}

Figure~\ref{fig:dist} shows distributions of the four indicators with target reference lines; Figure~\ref{fig:box} shows scenario boxplots of carbon-intensity reduction and water saving.

\subsection{Target-Attainment Probabilities}

Using the platform green targets (pesticide $\geq$20\%, fertilizer $\geq$15\%, aggregate water saving $\geq$20\%) and a carbon-intensity decline threshold ($\geq$15\%), attainment probabilities are: pesticide 73.8\%, fertilizer 90.6\%, aggregate water saving only 19.9\%, and carbon-intensity $\geq$15\% 98.1\% (carbon $\geq$10\% near 100\%). The low water-saving attainment mainly reflects the limited saving potential of the rice scenario (median 11.2\%) and the adoption discount; mango drip irrigation reaches a median saving close to 20\% (19.7\%), supporting scenario-specific target setting.

\subsection{Sensitivity Analysis}

\begin{figure}[htbp]
\centering
\includegraphics[width=0.75\textwidth]{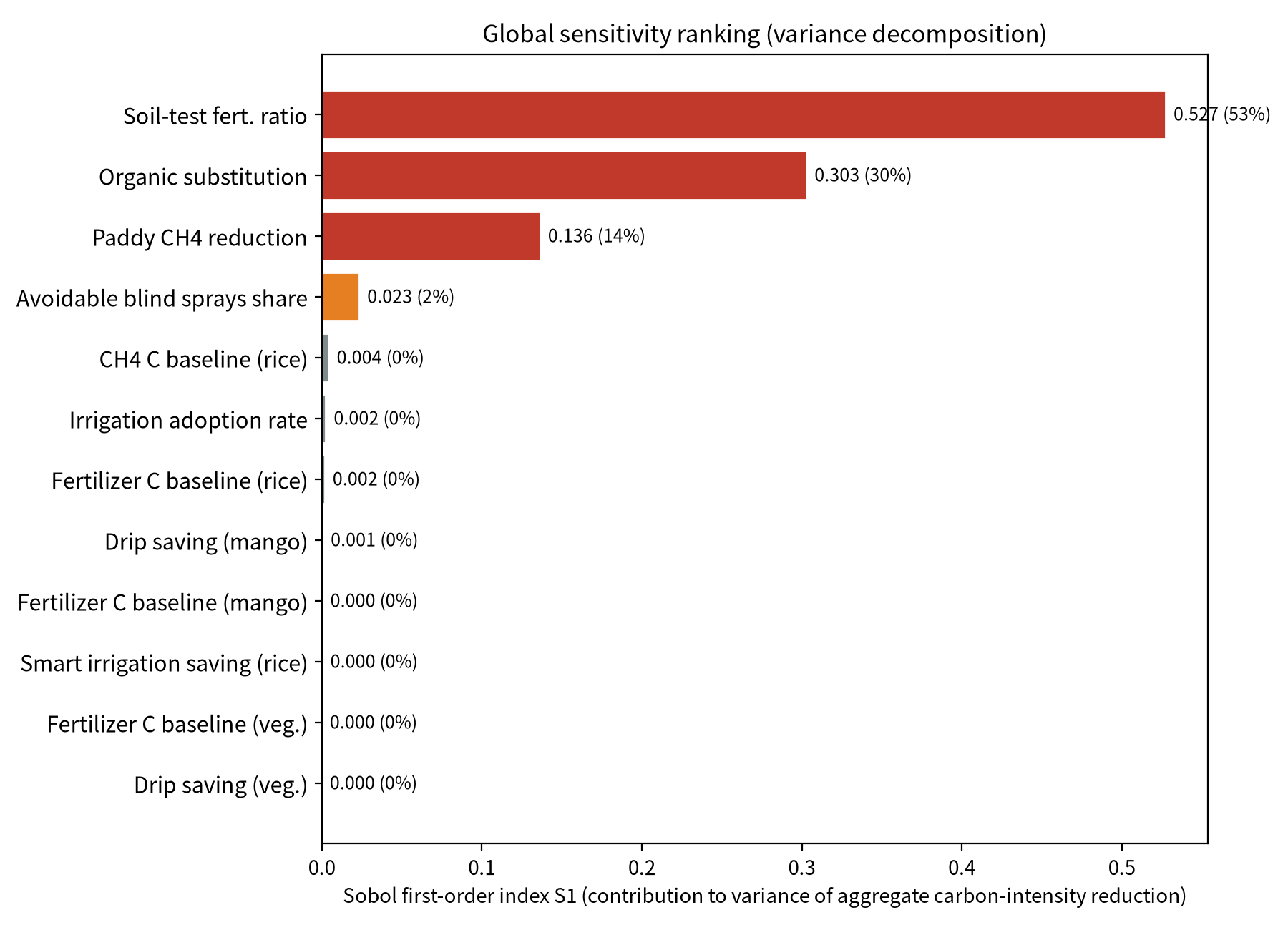}
\caption{Sobol first-order indices for aggregate carbon-intensity reduction (bracketed values are relative contribution shares)}
\label{fig:sobol}
\end{figure}

Sobol first-order indices for aggregate carbon-intensity reduction rank as: soil-test reduction $S_1=0.527$, organic substitution $S_1=0.303$, paddy CH$_4$ reduction $S_1=0.136$; the remaining parameters jointly contribute less than 5\% (Figure~\ref{fig:sobol}). The total S1 near 1.0 indicates a main-effect-dominated model with weak interactions. Implication: prioritizing pilot observation of nitrogen application and soil-test data, organic-substitution ratio and paddy water-layer management execution would capture most of the uncertainty in green value.

\subsection{Convergence and Scenario Checks}

\begin{figure}[htbp]
\centering
\includegraphics[width=0.68\textwidth]{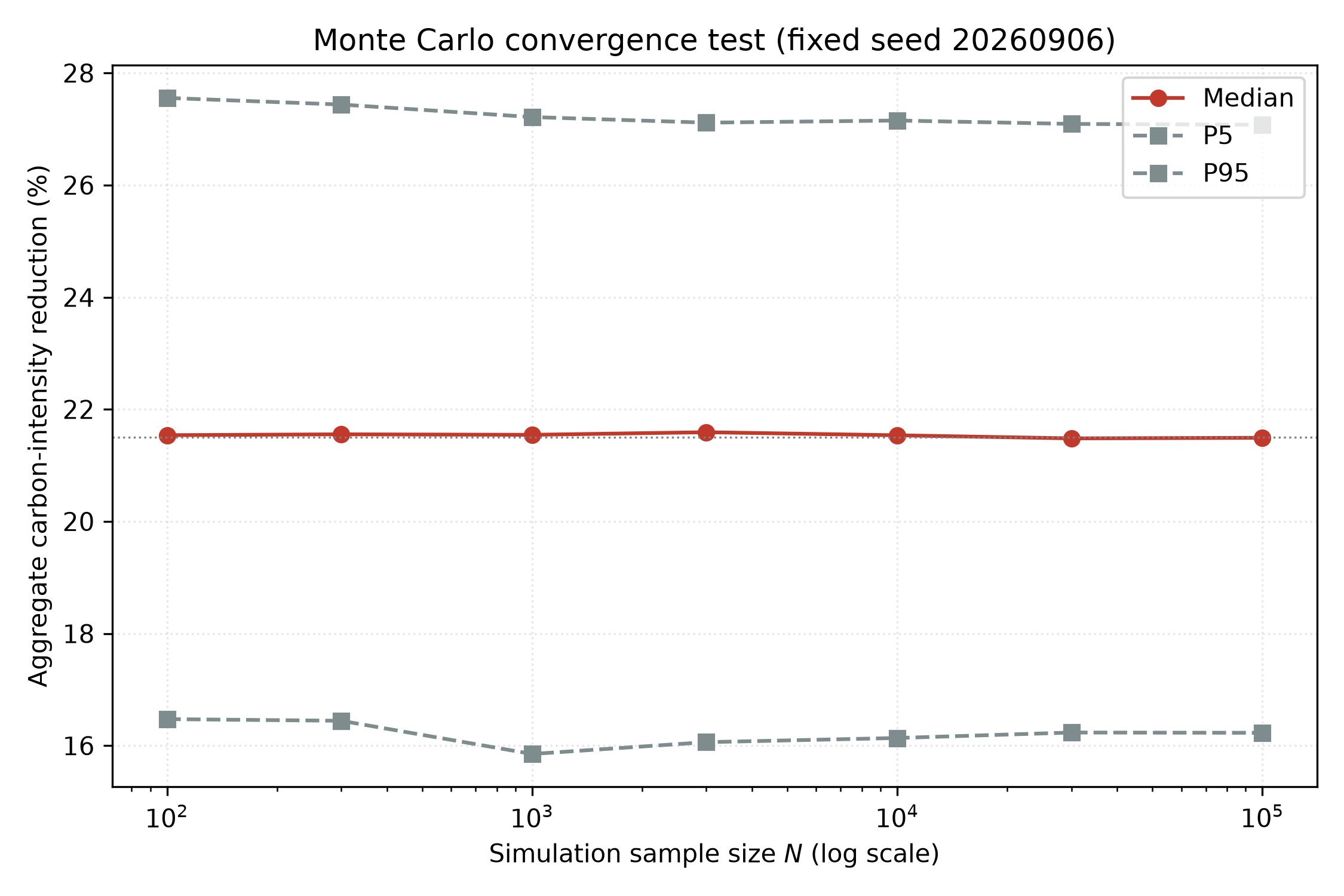}
\caption{Monte Carlo convergence: aggregate carbon-intensity reduction statistics vs.\ sample size}
\label{fig:conv}
\end{figure}

Convergence tests (Figure~\ref{fig:conv}) show that beyond 1,000 samples the median aggregate carbon-intensity reduction stabilizes around 21.5\% (P5/P95 around 16.2\%/27.1\%); statistical error at 10,000 iterations is negligible.

The three-scenario analysis (Table~\ref{tab:scenario_res}) yields conservative--baseline--optimistic ranges: pesticide reduction 11.3\%--37.5\%, fertilizer reduction 6.5\%--37.5\%, aggregate water saving 4.5\%--38.0\%, aggregate carbon-intensity reduction 6.9\%--38.8\%. Even under the most unfavorable parameter combination, carbon-intensity reduction remains positive (6.9\%), so the directional conclusion of a clear carbon decline is robust; however, at lower bounds of adoption and parameters, pesticide/fertilizer/water indicators fall clearly below design targets, highlighting the key role of adoption rate and field-level implementation.

\begin{table}[htbp]
\centering
\caption{Three-scenario analysis (parameter endpoints)}\label{tab:scenario_res}
\small
\begin{threeparttable}
\begin{tabular}{lcccc}
\toprule
Scenario & Pesticide & Fertilizer & Agg. water & Carbon\\
 & reduction & reduction & saving & intensity red.\\
\midrule
Conservative (lower bound) & 0.113 & 0.065 & 0.045 & 0.069\\
Baseline (mode) & 0.225 & 0.195 & 0.160 & 0.200\\
Optimistic (upper bound) & 0.375 & 0.375 & 0.380 & 0.388\\
\bottomrule
\end{tabular}
\end{threeparttable}
\end{table}

\subsection{Comparison with Literature-Reported Values}

Medians are compared with literature/official ranges in Table~\ref{tab:lit}. Apart from irrigation water saving (this study includes the adoption discount and is therefore below the official upper bound), all medians fall inside literature ranges, supporting the plausibility of the model settings.

\begin{table}[htbp]
\centering
\caption{Comparison of simulated medians with literature/official values}\label{tab:lit}
\small
\begin{threeparttable}
\setlength{\tabcolsep}{3pt}
\begin{tabular}{llll}
\toprule
Indicator & This study median & Literature range & Reference/basis\\
\midrule
Pesticide reduction & 0.235 & 0.15--0.30 & IPM literature\\
Fertilizer reduction (N) & 0.210 & 0.10--0.25 & soil-test evaluation\\
Irrigation water saving & 0.197 (mango) & 0.20--0.50 & water-fertilizer guidance \cite{moa2013}\\
Paddy CH$_4$ reduction & 0.300 & 0.20--0.50 & IPCC 2019 water mgmt. \cite{IPCC2019}\\
Carbon-intensity reduction & 0.215 & 0.15--0.35 & precision-ag. reviews\\
\bottomrule
\end{tabular}
\end{threeparttable}
\end{table}

\section{Discussion}

\subsection{Methodological Discussion}

The key difference between this framework and earlier single-technology assessments is that all platform modules (question answering, diagnosis, sensing, remote sensing, closed-loop records) are uniformly encoded as quantifiable transmission parameters, allowing benefit intervals at the whole-platform level. The approach inherits the mature paradigm of LCA with Monte Carlo uncertainty propagation \cite{saltelli2008,lam2014}, while avoiding hard dependence on field-experiment data, which suits early-stage assessment.

Uncertainty sources are decomposed into three classes: prior parameter uncertainty (grade-C parameters), structural uncertainty (different transmission-chain configurations) and scenario uncertainty (adoption rate). They are characterized here by triangular distributions, Sobol decomposition and three-scenario analysis. Grade-C parameters (e.g., blind-spray share 0.15--0.50) have wide ranges and are the main source of uncertainty for pesticide results; pilot observation should first collect local Hainan data to tighten them.

\subsection{Policy and Application Implications}

(1) "Fertilizer reduction $\geq$15\%" and "clear carbon-intensity decline ($\approx$21\%)" are the two most credible green indicators and can be prioritized in target assessment; (2) "aggregate water saving 20\%" has only about 20\% attainment probability, so scenario-disaggregated statements are preferable---mango drip and vegetable water-fertilizer scenarios show median saving of 17\%--20\%, while the rice scenario is limited; the aggregate is more robustly expressed as a 10\%--25\% interval; (3) paddy water management (intermittent irrigation) offers additional CH$_4$ mitigation potential in Hainan's double-cropped rice and nanfan fields (contributing about 14\% of the variance of carbon-intensity reduction), a low-cost high-benefit green lever.

\subsection{Limitations and Outlook}

Limitations: (1) this is a conditional full-adoption ex-ante scenario that does not model learning curves, seasonal differences or partial adoption; (2) yield effects are not considered---precision management usually maintains or slightly raises yield, so not accounting for yield-scaled intensity makes results conservative; (3) carbon baselines rely on literature/databases, and different inventories (Ecoinvent vs.\ Chinese product carbon-footprint factors) can differ by a factor of two for pesticide footprints; relative reduction rates are therefore recommended as primary indicators; (4) the paddy CH$_4$ default factor depends strongly on soil organic matter, straw return and water regime and needs Hainan field validation. Future work will: (1) observe the four indicators in pilot plots and update parameter posteriors by Bayesian methods; (2) introduce seasonal and adoption dynamics models; (3) extend accounting to post-harvest stages and ecological product value pathways.

\section{Conclusions}

(1) Under full-adoption simulation, the median green benefits of the platform are: pesticide reduction 23.5\% (90\% interval 15.0\%--33.2\%), fertilizer reduction 21.0\% (13.8\%--28.9\%), aggregate irrigation water saving 16.5\% (10.9\%--23.5\%), and carbon-intensity reduction 21.5\% (16.1\%--27.2\%).

(2) Attainment probabilities of "fertilizer reduction $\geq$15\%" and "carbon-intensity decline $\geq$15\%" are about 91\% and 98\%, respectively, with high credibility; "aggregate water saving $\geq$20\%" has only about 20\% attainment probability and should be reported on a mango/vegetable drip scenario basis.

(3) Sobol analysis shows soil-test rate and organic-substitution coefficient contribute about 83\% of the variance of aggregate carbon-intensity reduction; pilot observation should prioritize fertilizer-input and paddy water-layer indicators.

(4) The Monte Carlo framework combined with convergence and three-scenario checks provides a reproducible, calibration-ready methodological reference for green-value assessment of smart agriculture platforms.

\section*{Data and Code Availability}
All simulation scripts and sample data are available in the delivery directories (scripts/, output/) with fixed random seed 20260906; all results in this paper can be reproduced in one command run.

\section*{Acknowledgements}
The authors thank the Sanya University innovation and entrepreneurship incubation platform, Handan Weizhi Jingjie AI Basic Software Co., Ltd., and the agricultural expert advisory team for their support in knowledge-base construction and parameter-range definition. This work was supported by the National College Student Innovation and Entrepreneurship Training Program (China).

\renewcommand{\refname}{References}
\bibliography{refs}

\end{document}